# Defining data science: a new field of inquiry

Michael L. Brodie mlbrodie@seas.harvard.edu
Data Systems Laboratory, School of Engineering and Applied Sciences
Harvard University, Cambridge, MA USA

=============DRAFT July 12, 2023 =====================

Data science is not a science. It is a research paradigm. Due its power, scope, and scale, it will surpass science – our most powerful research paradigm – in enabling knowledge discovery, widely predicted to change our world[12]. We have yet to understand and define it.

Modern data science is in its infancy. Emerging slowly since 1962 and rapidly since 2000, data science is a fundamentally new field of inquiry, one of the most active[1], powerful, and rapidly evolving innovations of the 21st century. Due to its value, power, and scope of applicability, it is emerging in over 40 disciplines, hundreds of research areas, and tens of thousands of applications. Yet we are just beginning to understand and define it. $10^6$ data science publications contain myriad definitions of data science and data science problem solving. Due to its infancy, many definitions are independent, application-specific, mutually incomplete, redundant, or inconsistent, hence so is data science as a field of inquiry. This research addresses this *data science multiple definitions challenge* by proposing the development of coherent, unified definition of data science based on a data science reference framework[25]-[31] and a data science journal[31] for the data science community to achieve such a definition.

This paper provides candidate definitions for essential data science artifacts that are required to discuss such a definition. They are based on the classical *research paradigm* concept[15] consisting of a *philosophy of data science*, the *data science problem solving paradigm*, and the six component *data science reference framework* – axiology, ontology, epistemology, methodology, methods, and technology that is a unifying framework that is frequently called for[1][4][7][10][11][16][19] with which to define, unify, and evolve data science. It presents challenges of defining data science, solution approaches, i.e., means for defining data science, and their requirements and benefits – the basis of a comprehensive solution [24]-[32].

## 1. Challenges defining data science

### 1.1. Unifying multiple definitions

The emergence of a new field of inquiry and its problem solving techniques is rare. Science and modern scientific analyses emerged 400 years ago and interpretivism and interpretivist analysis 200 years ago. While conventional data science is as old as mathematics, AI-based data science is in its infancy. Tukey's 1962 vision of exploratory data analysis[20][21] brought renewed attention to statistic's learning and discovery power. After 2000, machine learning-based data science led to a fundamentally new, inscrutable field of inquiry that we are just beginning to understand. Valuable contributions – concepts, methods, solutions – from hundreds of disciplines and thousands of applications, while adequate for their source area must be unified to apply across data science and in thousands of applications. This has led to calls for a unifying framework to guide unification. An analogous problem arose in science in the 1830s.

What is such a unifying framework? How do you define a fundamentally new field of inquiry? For this we look to science, our currently most powerful knowledge discovery paradigm.

### 1.2. Classical paradigms to define science

As described in Section 2, data science is a research (knowledge discovery) paradigm distinct from and independent of science – our most mature, successful, and powerful research paradigm. 2,600 years ago science was defined philosophically by Thales of Miletus (624-560 BC) and Aristotle (384-322 BC)) then in terms of scientific models, theories, and the scientific method by Francis Bacon [Novum Organum 1620] and David Hume (1739-1779)[5] and only recently (1962, 1970) was the philosophy of science defined by Thomas Kuhn[13]. Consider the following classical paradigms used to define and unify science to be used to understand, define, and unify data science. Knowledge of science contributes to understanding data science by contrast.

---

[1] 500,000 AI articles were published in 2021, more than in any other discipline, dominated by pattern recognition, machine learning, and computer vision[22]. ACM lists 200+ data science journals.

Just as data science is now enthusiastically adopted and applied worldwide, so it was with science in the 19th century. In the 1830s, science was emerging in many disciplines and faced the multiple definitions challenge for science, scientific problem solving, and scientific disciplines. Scientific research and analyses (experiments) emerged rapidly, simultaneously, and independently in many universities, industries, and countries, resulting in many publications, each with its own definitions and terminology. Different definitions emerged from different scientific experiments and disciplines, from scientists with different backgrounds and knowledge, and in different discipline-specific publications.

In 1837, Alexander Comte[15] recognized the multiple definitions challenge for science as chaotic due to science and scientific disciplines being inadequately understood and defined. Definitions were adequate for experiments, or scientific problem solving in source scientific disciplines but were often inadequate for scientific problem solving across all scientific disciplines. Myriad independent scientific definitions were incomplete, mutually inconsistent and redundant with inadequate guidance to understand and conduct science in as yet undefined scientific disciplines. Myriad definitions impeded scientific understanding, communication, and collaboration, – a persistent problem to this day[8]. A comprehensive solution to the multiple definitions challenge for science and scientific disciplines was to deduce a unified definition of science and scientific disciplines from myriad, independent definitions. This required paradigms that were accepted by scientists to guide the unification of the myriad definitions based on established results. Comte provided those paradigms. This research addresses the analogous data science multiple definitions challenge and follows Comte's model used for science. The solution in science took 200 years. Defining and unifying data science may take decades due to its current inscrutability.

Comte introduced two paradigms and the concept of a discipline that have been used ever since[15]. He defined the *scientific research paradigm* in terms of *the philosophy of science* and a *scientific reference framework*. The philosophy of science is a worldview that provides the philosophical underpinnings (i.e., objective, quantitative reasoning) of empirical research for knowledge discovery, with which to understand, reason about, discover, articulate, and validate *the true nature of the ultimate questions about natural phenomena* as new knowledge about the phenomena. Scientific results are definitive, conclusive, casual, robust, universal knowledge about the phenomena with verified explanations and validated interpretations that are not provably complete.

The scientific reference framework defines science in terms of its axiology, ontology, epistemology, methodology, methods, and technology. Observing that scientific discovery is conducted only on specific phenomena within a discipline, he defined a *scientific discipline* as the scientific research paradigm applied to a specific class of phenomena, e.g., biology is the application of the scientific research paradigm to living organisms. Disciplines can be structured into sub-disciplines for sub-categories of a discipline's phenomenon, e.g., microbiology is a sub-discipline of biology that deals with a sub-category of living organisms, i.e., macromolecules that are essential to life. Surprisingly, the concept of a scientific discipline did not exist until Comte's definition. Observing that the central purpose of the scientific research paradigm is scientific problem solving, Comte defined the *scientific problem solving paradigm* governed by the scientific method. Comte's definitions of the scientific research paradigm, a scientific reference framework, a scientific discipline, the scientific problem solving paradigm, and discipline-specific scientific problem solving provided a solution to the multiple definitions challenge in science. Comte's 1830's paradigms defined modern science and guided its conduct in scientific disciplines ever since. The *research paradigm approach* that applies the above paradigms to define science is now used to define data science for which candidate definitions are proposed for essential data science concepts. The essential concepts are then used to address the multiple definitions challenge for data science.

**1.3. Classical paradigms to define data science**

A definition of data science is of value to the extent that it meets the needs of and is accepted by the data science community. The *data science community* consists of data science researchers, developers, practitioners, educators, and students that research, develop, and apply data science in 40 contributing disciplines, hundreds of research areas, and tens of thousands of applications in an ever increasing number of data science disciplines. We need at least candidate data science definitions to discuss such an activity. This section provides informal, candidate definitions of essential data science concepts and problem solving as a starting point for the data science community using an online data science journal[32].

*Paradigm* – a well-defined concept, model, or template – is used to understand and define data science. "In science and philosophy, a paradigm is a distinct set of concepts or thought patterns, including theories, research methods, postulates, and standards for what constitute legitimate contributions to a field" [Wikipedia] Levels of abstraction are used extensively. A paradigm can have many distinct instances, i.e., paradigm categories[2]. Comte was concerned exclusively with science; hence, his scientific research paradigm was actually a research paradigm category as is data science ; however, science and data science are two of over 30 such research paradigms. Paradigm categories have many instances, e.g., the scientific research paradigm category[3] has many scientific disciplines defined as instances of applying the scientific research paradigm category to a specific class of phenomena. Levels of abstraction can be confusing as the precision required for definitions leads to lengthy terms, e.g., the full term *data science research paradigm category discipline* is required for precision; when the meaning is clear, the simpler term *data science discipline* is used. Similarly, when the meaning is clear, the simple term *data science reference framework* is used in place of longer *data science research paradigm category reference framework*.

Research paradigms used in this research were introduced indirectly and perhaps unwittingly by Comte. Comte tried to apply science, called positivism at the time, to knowledge discovery in sociology. "Although the positivist approach has been a recurrent theme in the history of western thought, modern positivism was first articulated in the early 19th century by Auguste Comte. His school of sociological positivism holds that society, like the physical world, operates according to general laws. After Comte, positivist schools arose in logic, psychology, economics, historiography, and other fields of thought." [Wikipedia] As with today's enthusiasm for applying data science extensively, so it was with science in the 18th and 19th centuries. Comte believed in the value of science and of scientific problem solving in humanistic topics. This led science to be applied inappropriately to disciplines for which the scientific method, objectivity, and quantitative reasoning were impossible and subjective, qualitative evaluations were essential. In the late 19th century, the positivist paradigm was replaced for such disciplines with the *interpretivist paradigm category* – a fundamentally new research paradigm independent of science with a philosophy of subjective, qualitative reasoning based on human knowledge, expertise, experience, biases, and opinions. Interpretivism governs disciplines in the arts and humanities. Similar confusions now exist when data science is considered as a science.

---

[2] The term category is used here to refer to a paradigm instance.

[3] *Scientific research paradigm* is an imprecise term for the *scientific research paradigm category.*

Comte's definitions with references to science removed define the generic research paradigm, research paradigm discipline, and research paradigm problem solving paradigm, as follows. The (generic) *research paradigm* is defined by its philosophy, and its reference framework[15][17]. The *philosophy of a research paradigm* is a worldview that defines the purpose of work in the research paradigm and its nature, i.e., how work is conducted, e.g., Appendix I. The philosophy provides the philosophical underpinning for research with which to discover, reason about, understand, articulate, and validate *the true nature of the ultimate questions about phenomena in a specific discipline* as knowledge about those phenomena. The *research paradigm reference framework*[4] consists of six components – axiology, ontology, epistemology, methodology, methods, and technology. Full definitions of the components can fill entire books. They are defined here briefly to understand and define data science. *Axiology* defines the purpose, nature, importance, risks, and value of the research paradigm for reasoning, problem solving, and the problem solving results, i.e., discovered knowledge. An o*ntology* and an *epistemology* are to a research paradigm what footings are to a house; they form the conceptual foundations of the edifice. For a research paradigm, the foundations are its artifacts (ontology) and means for reasoning with those artifacts (epistemology). Together they constitute a vocabulary and language of the research paradigm. *Methodology* defines the process, or methodology, referred to as *the method* of the research paradigm, for defining, designing, and conducting analyses (work), their expression in a workflow, and the governing principles, e.g., the scientific method governs the design and execution of scientific analyses expressed in scientific workflows governed by scientific principles. *Methods* is the heart of research paradigm analyses – analytical means used to implement analyses in the research paradigm defined by analytical methods, operands, solutions (prepared models and operands), results, and governing principles. *Technology* defines technical and engineering concepts, tools, languages, platforms, and systems governed by engineering principles and practices used in practice to implement artifacts and analyses in the research paradigm. The definition of each component is complex involving many artifacts (concepts, models, methods, analyses, principles, and laws), reasoning, and technology systems.

The definition of the research paradigm reference framework is specialized to define specific research paradigms, e.g., science and data science, that in turn are specialized to define its disciplines within the research paradigm. This simple yet powerful rule, the *paradigm approach*, central in this research, is described below.

The problem solving paradigm has been used for centuries to analyze (solve) complex domain problems that are expressed in terms of a specific class of phenomena. The *problem solving paradigm* of a research paradigm is defined in its reference framework. The methodology component defines the process that governs defining, designing, and executing a solution and its expression in workflow steps. The methods component defines the analytical methods (means), their operands, the solutions and results that are used in workflow steps to conduct problem solving. The following is a detailed description of the (generic) problem solving paradigm in a research paradigm, i.e., the process[5] that governs the definitions, design, and execution of an analysis as a solution for a domain problem. This description does not address the frequent

---

[4] For modern science and data science, methods and technology are added to Comte's definition.

[5] A research paradigm, e.g., science, has one methodology, e.g., the scientific method governs the design and execution of experiments; analyses in data science are governed by *the data science method.*

case of a domain problem that consists of multiple subproblems.

The objective of generic problem solving paradigm (Fig. P2) for a motivating domain problem is to design a domain solution with an explanation that verifies that it solves the intended problem and produces a domain result with an interpretation that validates the result is the intended domain result. Domain problem solving is done in terms of instances of a specific class of phenomena. To aid domain problem solving, the domain problem is translated to an equivalent analytical problem expressed in terms of analytical models and methods used to design and execute an analytical solution to produce an analytical result with an explanation that verifies the analytical solution and an interpretation that validates the analytical result. Analytical problem solving is done on data – models of the phenomenon being analyzed and data analyses conducted over that data.

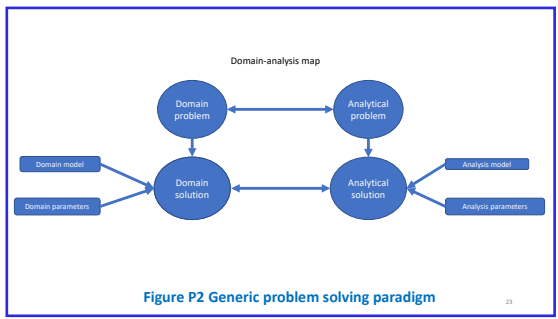

Figure P2 Generic problem solving paradigm

The purpose of analytical problem solving is to use data and data analysis to provide insights into the corresponding domain problem, solution, and result. The benefits include providing problem solving techniques (models, methods) to augment those on the domain side and to overcome limitations of domain side models and methods. Particle physics experiments are not possible without analytical models and methods like the standard model of particle physics. Analytical problem solving requires demonstrating the equivalence of the domain and analytical sides.

The problem solving method for the generic problem solving paradigm involves several steps on both sides. On the domain side: Step 1 Use established knowledge, or model, of the domain phenomenon to define the domain problem in terms of that model and methods. Step 2 Use domain-specific problem solving means (e.g., a particle physics cyclotron) to develop a domain solution. Step 3 Develop a solution explanation to verify the solution. Step 4 Use the domain solution to produce a domain result. Step 5. Develop an interpretation of the domain result to validate that it solves the domain problem. On the analytical side, for each analytical step attempt to establish the equivalence with its domain counterpart: Step 1 Use established analytical knowledge, or model, of the domain phenomenon (e.g., observed evidence of the Higgs boson) and the domain problem definition to define the analytical problem in terms of an analytical model. Step 2 Use domain-specific analytical problem solving means to develop an analytical solution (e.g., particle physics simulator). Step 3 Develop a solution explanation to verify the solution. Step 4 Use the analytical solution to produce an analytical result. Step 5. Develop an interpretation of the analytical result to validate that the analytical result solves the analytical problem. While step 1 on both sides are done first, the order of steps on each side and interleaving between sides is driven by need. A step on one side may assist a step on the other, e.g., domain steps provide insight into its analytical counterpart and *vice versa*.

In complex problem solving, domain and analytical problems are not solved in isolation. More context is required. For example, an experiment cannot detect a Higgs boson in isolation. A 1.25GeV energy event required to detect a Higgs boson is not an isolated event. It produces an energy cascade in which a Higgs boson exists momentarily before decaying into leptons and other particles. For complex problem solving, the generic problem solving paradigm is augmented to the generic problem solving paradigm (Fig P3) with two additional levels, the model and the standard solution levels. The domain and analytical problems (Fig P2) each exist in a larger context called the domain and analytical models, respectively with additional knowledge to assist problem definition and solving. For example, detecting the Higgs boson requires detecting its signature energy cascade. The model of the atom and its behavior is the domain model to better define the domain problem. The standard model of particle physics is the analytical model to better define the analytical problem. A successful domain result becomes new knowledge about the phenomenon that is curated into the domain model. For example, the signature Higgs boson cascade was detected, hence the Higgs boson was detected. The existence of the Higgs boson and its signature energy cascade is curated into the domain model of the atom and into the analytical model – the standard model of particle physics.

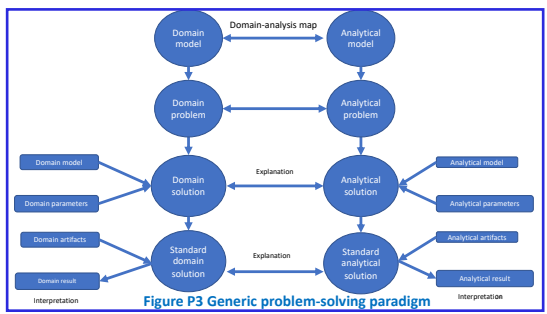

Figure P3 Generic problem-solving paradigm

While there is considerable flexibility in the order of the steps in a problem solution, the steps and their order for a specific solution is expressed in a problem solving workflow that consisting of three phases. The *Discover phase* defines a problem then designs, implements, producing a result (steps 1, 2, 4), and analyzes the solution and result. The *Interpret phase* articulates, confirms, authorizes, and curates the solution and result (steps 3, 5). The *Deploy phase* deploys the solution. Each step is implemented using one or more methods defined in the methods component of the research paradigm reference framework. The workflow and the methods are governed by principles of the research paradigm, e.g., repeatability in the scientific method.

A *research paradigm (category)* is defined at two levels of abstraction. At the abstract (generic) level, a research paradigm is defined by its artifacts, defined in its ontology component, and by its properties, capabilities, and limitations that are, in part, deduced collectively from values of artifact instances at the lower, applied level. A paradigm has a value for each research paradigm reference framework artifact. The two levels uniquely define the research paradigm, distinguishing it from all other research paradigms. For example, the scientific research paradigm discipline biology is defined in terms of generic artifacts of its phenomena, living organisms (e.g., name, taxonomic ID (domain, kingdom, phylum, class, order, family, genus, and species), digestive system, form of mobility, etc.). Each biology sub-discipline has a specific range of values for each artifact instance. The unique properties, capabilities, and limitations of each biology sub-discipline are derived, in part, from its artifacts' values. For example, the living organism subcategory cheetah is the fastest land animal due, in part, to its form of mobility.

A *discipline within a research paradigm* is defined by applying the research paradigm to a class of phenomena. For example, a *data science discipline* is the application of the data science research paradigm to a class phenomenon, e.g., the data science NLP discipline is the data science research paradigm applied to natural language problems. Defining a *research paradigm discipline* is complex. It is done by specializing each research paradigm reference framework component and problem solving paradigm to the discipline's class of phenomena. The particle physics discipline is defined by applying the scientific research paradigm to the class of elementary particle phenomena. This requires a deep understanding of the scientific research paradigm and of the participle physics domain. This research provides guidance for defining data science disciplines but does not define any such disciplines that require the definition of all components of their data science discipline reference framework.

The *problem solving paradigm* of a research paradigm defines how work is conducted in a discipline. Its definition is similarly complex. Work includes[6] reasoning (discover, analyze), articulating (explain, interpret), confirming (verify, validate), authorizing, and curating (unify, organize) knowledge about phenomena in that discipline within a research paradigm. That work, the central purpose of a research paradigm, is defined by the methodology and methods components of the research paradigm's reference framework. The methodology component defines the process for defining, designing, and conducting analyses, i.e., developing solutions, the workflow for expressing a solution, and their governing principles. A problem solving paradigm has one methodology paradigm and one workflow paradigm with a vast number of categories, e.g., the scientific method and the scientific workflow in science, and the data science method[7] and data science workflow in data science. The methods component defines the analytical methods, their operands, solutions, results, and governing principles used in workflow activities. It defines the nature of *solution explanations* that explain how the problem was solved and is used to understand and verify that the intended analysis was conducted successfully; and of *analytical result interpretations* that interpret the result in terms of the analytical problem, and, more importantly, the motivating domain problem, and is used to understand and validate those results.

---
[6] Appendix I lists reasoning research paradigm actions (verbs).

**1.4. Definition challenges**

The above classical paradigms, categories, and disciplines are used to address the data science multiple definitions challenge by focusing on the two central concepts – data analysis artifacts and data analyses – each at three levels of abstraction, resulting in six definitional challenges posed as:

1. Data science: *What is it?* What is data science as a research paradigm category, or field of inquiry? How do you define a fundamentally new field of inquiry?
2. Data science discipline: *How is it used?* What is a data science (research paradigm category) discipline? How do you a define a data science discipline across all data science disciplines?
3. A distinct data science discipline: *How is data science applied in a specific data science discipline?* What constitutes a specific data science discipline, e.g., data science NLP? Is there value in unifying multiple definitions for data science NLP?
4. Data analyses: *What are they?* What are data analyses as a problem solving paradigm category? How do you define a fundamentally new problem solving paradigm category?
5. Data analyses conducted in a data science discipline: *How is it used?* What constitutes the data science problem solving paradigm category as conducted in a data science discipline? How do you define data science problem solving across all data science disciplines?
6. Data science problem solving as conducted in a specific data science discipline: *How is it applied in a specific data science discipline?* What constitutes a data science problem solving paradigm category in a specific data science discipline? Is there value in unifying multiple definitions of data science NLP problem solving?

The six cases are amenable to the same approaches, processes, and means to achieve solutions – unified, coherent, evolving definitions. Understanding and solving one aids understanding and solving others. As data science is in its infancy and is evolving rapidly, definitions are intended to capture the nature of artifacts, e.g., the nature of data science problem solving and of the resulting data science solutions, with flexibility to support inexorable evolution.

The myriad data science definitions, accepted through proofs of utility in their respective communities, may lead to confusion due to the multiple definitions challenge hence limit communication and collaboration[8] across data science disciplines, hence across data science as a field of inquiry. Data science has reached a level of maturity as seen in its acceptance by the large data science community to initiate unifying data science to overcome the multiple definitions challenges guided by Comte's paradigms.

---

[7] ***The*** data science method is to data science as the scientific method is to science; it governs the expression and conduct of data analyses in workflows. ***A*** data science method is a computational analysis category that can be implemented by one or more algorithms. This dual use of the term "method" arises in conventional use in research paradigms and in computations.

**1.5.　Definition solution approaches**

This research proposes a solution for defining data science and addressing the multiple definitions challenge, answering the six problems, using the research paradigm approach augmented by three additional approaches. Results of applying the solution are illustrated by 1,000 candidate definitions in [24]-[32].

The process for developing unified, coherent definitions and means to express them applies to all six cases. It is described only for the research paradigm category (i.e., What is data science). Step 1: Collect all relevant definitions of the data science research paradigm.[8] Step 2: Use the research paradigm definition to align their respective artifacts. Step 3: For each artifact, derive (coalesce) a unified understanding from the aligned artifact definitions. Step 4: For each artifact, use its unified understanding together with its research paradigm definition, i.e., as a unifying framework, to derive a unified, definition consistent with the research paradigm definition. Step 5: Use a specification language to express a unified, coherent definition. Step 6: For each source definition, examine the source to identify and resolve challenges that arise from changing the definition. If possible, make appropriate source changes. Step 7: Use the unified definition of the data science research paradigm to determine if it is a new, independent research paradigm category. Each of these steps can be complex. Requirements and benefits are provided for each problem.

To appreciate the process's scale and scope, consider the corresponding case for science. The scale of the process for science can be estimated by the number of artifacts defined in a dictionary of science, on average 20,000. The scope of such a process can be seen in the number of scientific disciplines and sub-disciplines since science grows by artifacts that emerge in existing and new scientific disciplines that without unification can lead to confusion[8]. There are ~16 major scientific disciplines including: chemistry, biology, physics, mechanics, computer science, psychology, optics, pharmacy, medicine, astronomy, archeology, economics, sociology, anthropology, linguistics. Biology has ~135 sub-disciplines [branches of biology UNC]. In practice, only the most accepted definition sources, e.g., journal publications, are considered in Step 1. Other complicating factors are problems that arise from differences between a unified artifact definition and source definitions. Changing a definition in a source context may impact related source definitions or negate established results. In most cases, changing source definitions is unlikely due to Kuhn's incommensurability thesis[13][20](§3.1). Alternatively, the unified definition may require modification or consider the unified definition as scientific progress and do not address consistency with source definitions.

To address the data science multiple definitions challenge within the above process, use three solution approaches – the first to understand, the second to define, and the third to manage coherence. The *paradigm approach* (steps 1-3) applies the data science research paradigm, defined in terms of generic data science artifacts and definitions of its categories defined by artifact instances. The research paradigm is a unifying framework used to align the artifact instances at the category level with their generic artifacts at the paradigm level. The alignment can lead to a unified understanding across the multiple definitions of data science research paradigm categories thereby to a unified understanding of the data science research paradigm and its categories. The *framework approach* (steps 4-7) uses the unified understanding of the data science research paradigm categories guided by the research paradigm definition as a unifying framework, to guide and structure a single, unified, coherent definition of emerging data science research paradigm with generic artifacts that are consistent with those in each data science research paradigm. The *semantic approach* is used in all steps to express and manage semantic relationships, e.g., generalization and composition (e.g., derived from, contains) amongst definitions to ensure coherence. The approaches work in two directions – deduction and derivation. Multiple definitions can be used to deduce a unified understanding and definition. An accepted paradigm definition can be used to derive multiple incomplete or inconsistent category definitions. The approaches can be taken based on definitions being mature, i.e., widely accepted within the data science community. They can be used to compare the resulting unified, coherent definitions of the data science research paradigm and its categories with definitions of the scientific research paradigm and its categories to establish that data science is a fundamentally new field of inquiry, independent of science as is done in §2.4.

---

[8] This example, posed in question 1, considers the data science research paradigm and its categories, I.e., data science disciplines. The process guides consistent definitions of the data analysis research paradigm and of data analysis disciplines.

Many research leaders [1][4][7][10][11][16][19] have called for a unifying framework with which to structure, develop, unify, and evolve data science definitions. "Data science without a *unifying framework* risk being a set of disparate computational activities in various scientific domains, rather than a coherent field of inquiry producing reliable reproducible knowledge. Without a flexible yet *unified overarching framework* we risk missing opportunities for discovering and addressing research issues within data science and training students in effective scientific methodologies for reliable and transparent data-enabled discovery"[19].

**1.6. Definition solutions**

The research paradigm, paradigm, framework, and semantic approaches are used to develop solutions to the data science multiple definitions challenge, i.e., to develop and express unified, coherent, evolving definitions for the six definitional problems.
1. *What is data science?* Data science (research paradigm category) is a field of inquiry, defined by applying the research paradigm approach to computational analysis over data.
2. *How is data science used?* Data science applies the data science problem solving paradigm to a category of phenomena in a data science discipline. There are as many data science disciplines as categories of phenomena that can be represented in data.
3. *How is data science applied in a specific discipline?* The data science problem solving paradigm applies computational means to analyze datasets that represent the discipline's phenomena, e.g., data science NLP applies neural network models for natural language problems.
4. *What is a data analysis?* A data analysis applies a trained, tuned computational method to dataset prepared to represent features of a phenomenon that are critical to the analysis. Data science problem solving across all disciplines is defined in terms of computational methods trained and tuned to analyze datasets prepared to be analyzed by a prepared method.
5. *How are data analyses used?* A data analysis selects, trains, and tunes a computational method, adding guardrails; acquires and prepares a dataset to be analyzed by that method resulting in a data analysis solution that can be applied to an unseen dataset to produce a data science result. The result provides insights to develop results for the motivating domain problem.
6. *How is a data analysis applied in a specific discipline?* A motivating domain problem in a discipline, e.g., protein folding, is expressed as one or more data science problems, e.g., 32 for AlphaFold. Each data science problem is solved to produce a data science solution (4 and 5 above) that may be combined, e.g., an ensemble solution for AlphaFold. The solution is applied to an unseen dataset, e.g., representing a specific protein, producing a result that provides insights for developing a domain solution, e.g., the 3D shape of the protein.

The resulting definitions can be used to define the emerging data science research paradigm and its disciplines as a fundamentally new, independent paradigm and its disciplines in terms of data science artifacts and data science analyses.

## 2. Science and data science are independent fields of inquiry

This section provides a definition of science as a field of inquiry based on Comte's paradigms for two reasons. First, it provides a detailed basis for comparing science and data science hence a deeper understanding of both. Second, the paradigms used to define science are used to define data science. The comparison shows that science and data science are distinct fields of inquiry, and that data science violates the definitive properties of science and *vice versa*.

### 2.1. The scientific field of inquiry

The scientific research paradigm is defined by the philosophy of science and the scientific reference framework (§1.2). The scientific ontology and epistemology components consist of definitions of scientific artifacts (concepts, models, methods, analyses, principles, and laws) and reasoning that comprise the language of science. The scientific methodology and methods components define the scientific problem solving paradigm, i.e., scientific analyses (experiments). The scientific methodology component defines the scientific method – how experiments are defined, designed, and executed, the scientific workflow – how experiments are expressed, and scientific principles that govern them, e.g., reproducibility, falsifiability, scientific induction. The scientific methods component defines scientific analytical methods (means), their operands (phenomena), solutions (experiments), and scientific results, e.g., a cyclotron is used to produce collisions of elementary particles (operands) to achieve particle cascades (results) in which to find evidence of the Higgs boson. The explanation of a scientific solution is the experimental design and its execution that explain how the analysis was designed and conducted permitting verification that the intended solution was achieved. The interpretation of a scientific result is inherent in the results of the self-explanatory cause-effect hypotheses thereby validating the scientific result.

In summary, a scientific analysis, conducted correctly following the scientific method and the principles of science and of the relevant scientific discipline, produces results that the hypotheses are either true or false. Scientific results are definitive, conclusive, casual, robust, and universal scientific knowledge with verified explanations and validated interpretations that are not provably complete. The power and elegance of science is that explanations and interpretations are inherent in the scientific method.

### 2.2. The data science field of inquiry

The *data science research paradigm* is defined by the philosophy of data science and the data science reference framework. The *philosophy of data science* is the worldview that provides the philosophical underpinnings (i.e., learning from data) for data science research for knowledge discovery with which to reason about (understand), discover, articulate, and validate[9] *insights into the true nature of the ultimate questions about a phenomenon by computational analyses of a dataset that represents features of interest of some subset of the population of the phenomenon.* Data science results are probabilistic, correlational, possibly fragile or specific to the analysis method or dataset, cannot be proven complete or correct, and lack explanations and interpretations for the motivating domain problem.

The *data science reference framework* consists of the data science axiology, ontology, epistemology, methodology, methods, and technology. A *data science ontology* consists of informal definitions of data science artifacts (concepts, models, methods, analyses, principles, and laws) and their relationships that define what can be expressed (represented) in data science. A *data science epistemology* provides means for reasoning over artifacts. Together they constitute the language of the data science. A data science problem is specified by the nature of analysis to be conducted over a dataset that represents the features of interest of the phenomenon to be analyzed. An ideal dataset represents the features that are critical to the analysis, of the entire population of the phenomenon, and has the highest entropy relative to the intended analysis. Such a dataset is impossible in data science since datasets are observational, i.e., not subject to specific constraints. A *data science solution* consists of a data science model trained, modified, and tuned to meet specific requirements to conduct the intended analysis on a data science dataset prepared for that analysis. It is considered a solution only after it has been modified, possibly with guardrails, to meet its requirements. A *data science result* is the computational result of executing the solution over a prepared dataset. The *data science problem solving paradigm* is defined by the methodology and methods components. The *data science methodology* component defines *the data science method*, the *data science workflow*, and principles that govern them. The *data science methods* component defines the constituents of data analyses – the *computational, analytical methods*, their *operands (datasets), solutions (trained models and prepared datasets), (computational) results*, and the governing principles. It also defines the nature of *data science solution explanations* and *data science result interpretation*. Due to the inscrutability of AI-based methods, the data science methods component cannot define explanations or interpretations but may provide techniques to address inscrutable solutions and results, e.g., *demonstrable, standard data science solutions* of which foundation models[3] are examples.

---

[9] Validation in science is inherent in the scientific method. Validation in data science is not currently possible, posing challenges for explanations and interpretations.

There are three categories of data science defined by computational methods. First, c*onventional data science methods* include mathematics, simulation, databases, data mining, statistics including probability theory and approximation theory, and some AI techniques like decision trees and linear SVMs. Such methods fall into well-known categories, e.g., clustering, outlier detection, association, classification, regression, summarization, factor, PCA, cohort, cluster, time series, sentiment, Monte Carlo simulation. While conventional data science methods have been used for centuries, only now are they recognized as the only transparent – scrutable – methods category, and the least powerful. In this well-understood category, solution explanations and results interpretations, while not inherent, are easier to construct than for AI-based methods, e.g., weather prediction models are designed, explained, and interpreted by experts using complex mathematics and simulations. The key factors here are that conventional methods and models are designed by humans to meet specific requirements hence humans are the agents of learning. The second and third categories are *AI-based methods* that emerged in the 1990s and now dominate data science. They are designed and conducted computationally by AI algorithms such as machine learning (ML), evolutionary, heuristic, and generative algorithms. Hence, the AI-based methods category is inscrutable lacking solution explanations and results interpretations. A key factor here is that the algorithm is the learning agent. That difference alone – human versus algorithmic design and learning – distinguishes AI-based data science from science. AI-based data methods has two subcategories *machine learning-based methods* and *deep learning-based methods. D*L-based methods, a sub-category ML-based methods, is a separate category as it is significantly more challenging to address. The three computational data science method categories are described further in[30].

A *data science discipline* is defined by applying the data science reference framework to a specific class of phenomena for which there is adequate data for data science analyses, e.g., data science NLP is a data science discipline that addresses NLP using large language models.

The data science method that governs the data science problem solving paradigm, illustrated in Fig. P4, involves four specializations of the generic

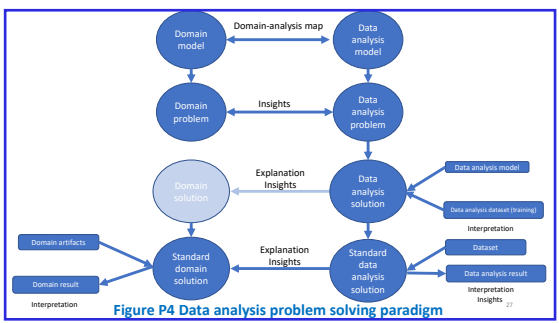
Figure P4 Data analysis problem solving paradigm

problem solving paradigm (Fig P3). First, data analyses take place entirely in data on the analytical side. Second, data science problem solving aids domain problem solving by providing insights into the motivating domain problem, solution, and result. Third, a domain problem is used to define corresponding data science problems but is rarely used to develop an often unsolvable domain solution. Data science solutions may provide insights into the domain solution for conventional data analyses but rarely do for AI-based data analyses. Fourth, increasingly classes of domain problems have been solved using data science. In such cases generic data science models, problem solutions, and results exist, e.g., in foundation models[3], and can be reused with techniques, e.g., transfer learning, to specialize them to specific domain problems. Similarly, foundation datasets can be used for specific features of specific phenomena.

Consider the data science problem solving paradigm for a domain model and problem concerning a specific phenomenon with no current data science solution or corresponding data science model and problem. The domain model and problem are developed as in the generic problem solving paradigm. Analytical problem solving are specialized for data science as follows.

First, the domain problem must be translated to a data science problem with their equivalence captured in a domain-analysis map to be maintained to assist later insights. Indirect approaches are common using one or more data science models and problems that map to parts of the domain model and problem. This challenging step requires deep knowledge on the domain and analytical sides. The DeepMind AlphaFold data science protein folding solution is a celebrated case. The protein folding problem was translated to 32 data science problems each with its own solution and results that were combined to form an ensemble analytical solution that can be applied to proteins to gain insights into the 3D structure of the protein. The AlphaFold team of over 40 of the world's leading experts in data science, genomics, and bioinformatics built on multiple existing data science models and standard data science solutions over several years and several previous models.

Second, a data science solution must be developed for the data science problem. Based on the nature of the analysis to be conducted, e.g., the analytical category and the nature of the available observational datasets, select the computational method and corresponding dataset that best represents the features of the phenomenon critical to the intended analysis. This step involves selecting, training, tuning, and evaluating multiple computational methods, acquiring and preparing multiple candidate datasets, and experimenting to select the optimal pair. A dataset that meets requirements can drive the selection of the data science method and *vice versa*. A computational method requires training and tuning for which there are many techniques, e.g., supervised, semi-supervised, unsupervised, and self-supervised, each with specific dataset requirements. Dataset preparation can involve partitioning into training, inference, validation, and verification datasets. As with the model, the dataset may require modification. Correcting errors in the dataset can improve the quality of the dataset and results. However, adding or eliminating data representing features of the phenomenon can have negative effects. By the data-driven principle of data analyses, eliminating data from the dataset eliminates any associated patterns from the results thus eliminates the corresponding insights into the phenomenon that they might have provided. The data science model is trained by computation over the dataset involving inference going forward from the initial state of the analytical model to the final state, called forward chaining, and backward chaining.

Forward and backward chaining can be repeated many times, e.g., thousands or millions, while being monitored and tuned or otherwise modified until the data science result meets specific requirements. This step creates a *data science solution*, i.e., a trained, tuned computational method for a prepared data science dataset that can be applied to a similar, unseen dataset to produce a *data science result*. An significant open challenge in developing a trained data science model is to constrain, using so-called guardrails, results to meet requirements, e.g., safety, trustworthiness, and prevent hallucinations.

Third, the data science solution and data science result are analyzed to develop an explanation of how it solves the intended data science problem and how to interpret the data science result in terms of the motivating domain problem. These are done to increase confidence of obtaining the correct solution and result and to be used as generic explanations and interpretations to be applied when the data science solution is deployed on an unseen dataset. This is easy for conventional data science methods but not for AI-based methods for which techniques such as demonstrable, standard data science solutions are used to aid in evaluating the uncertainty of the result and the risk of using the result to infer domain results to be used in practice. AI-based data science methods provide no technical solutions for these steps for which engineering techniques have been developed. Domain problem solutions[10] are seldom developed in such cases (shown as feint in Fig P4) as they are often beyond human capacity to understand in scope, scale, and complexity. This shows the strength of data science to develop data science results for such problems and its inability to develop domain solutions.

Fourth, a data science solution can be applied to an unseen dataset prepared for the analysis to produce a data science result. The fifth and final step is the most critical. The data science solution and result are used to develop insights into the domain problem, solution, and result to develop results for the motivating domain problem to be applied in practice.

The greatest risks posed by data science problem solving is the lack of technical solutions to control, understand, explain, or interpret data science solutions and results. Human designers, developers, and users are exclusively responsible for demonstrating that results are safe, reliable, transparent, explainable, interpretable, privacy-preserving, accountable, fair, trustworthy, reliable, and robust, often post-facto. Data science provides no technical means for this step.

The uncertainty of data science results poses risks in their use to develop results for motivating domain problems to be applied in practice. Data science solutions and results are inscrutable for several reasons. AI-based data science methods and their results are inscrutable. We do not understand what analysis is conducted, the certainty of the results, i.e., neither robust nor reliable, nor how to interpret them. The prepared datasets cannot be proven to accurately represent the features of the phenomenon most critical to the analysis nor the extent to which a dataset represents the entire population. Critical features could be missing or a feature pattern may not accurately represent its occurrence in the population. To be applied in practice, the result intuited for the motivating domain problem must be demonstrated to be within acceptable bounds. While data science solutions and results are inscrutable, many techniques have been and are being developed, often integrated into the problem solving process, to assist with developing interpretations with acceptable risks. For example, a well-defined and maintained domain-analysis map can assist in mapping data science results to domain results, as is done in data science model training techniques, e.g., supervised and self-supervised. Demonstrable, standard data science solutions, like foundation models, assist by developing and demonstrating probabilistic estimates of uncertainty and risks of applying a domain result in practice. More convincingly, means outside data science, e.g., empirical analysis of the behavior of a solution[11], are used. This illustrates the power and limits of data science to develop insights into motivating domain problems, solutions, and results.

---

[10] For most domain problems addressed with data science, the objective is to gain insights into the domain result with little or no interest in a corresponding domain solution.

Developing a *demonstrable, standard data science solution* addresses the most critical step five in data science problem solving – developing explanations and interpretations for the motivating domain problem within demonstrably acceptable bounds, as outlined below.

Domain problems concerning a phenomenon are posed in a model of the phenomenon based on established knowledge, called a standard model of the phenomenon, e.g., the standard model of particle physics for the Higgs boson problem and the central dogma of microbiology for the protein folding problem. Standard models provide a frame of reference within which domain problems are posed and solutions developed. The Higgs boson experiment was not discovered from scratch. It was discovered as the 61st elementary particle within a model consisting of 60 known elementary particles. The generic problem solving paradigm has standard models on the domain and analytical sides. The domain standard model is often used to guide the development of the analytical standard model and to establish a domain-analysis map.

This classical technique applies to data science problem solving. Apply the first step in data science problem solving using the domain standard model and problem to specify or select a corresponding data science standard model and problem and a domain-analysis map. As data science matures, there are more standard data science models, i.e., trained, tuned data science models for specific types of domain analyses on prepared datasets, e.g., foundation models and foundation datasets. While these uses of the domain-analysis map cannot be proven to be correct, they can be used to demonstrate that the selected data science model and dataset can be applied within acceptable bounds. Due to the inscrutability of AI-based data science methods, the map rarely aids in developing an explanation for the data science solution nor for its domain counterpart.

If there is an acceptable standard data science model and dataset and domain-analysis map, it may require modification for the specific application by applying steps two through four for additional model training (transfer learning), tuning, and refinement or addition of guardrails, and further dataset acquisition and preparation. If the results of step four are within bounds, then apply step five by modifying the generic explanations and interpretations of standard data science model and dataset to meet the needs of the specific domain problem and solution. If no acceptable standard data science solution and dataset can be found then previously mentioned engineering techniques may apply.

## 2.3. Data science and science are independent, complementary fields of inquiry

The unique properties, capabilities, and limitations of science are to seek by objective, quantitative means definitive outcomes for observable, i.e., measurable, bounded, phenomena versus data science that seeks by inscrutable means uncertain insights into phenomena beyond human capacity to understand in scope, scale, and complexity. As described in §2.2, data science results violate the principles of scientific results, and *vice versa*. Hence, data science and science are independent research paradigms. The contradictions between science and data science hold for the questions posed in §1.4 and solutions proposed in §1.6 .

These differences make science and data science powerful, complementary research paradigms. One of the most important 21st century experiments, the Higgs Boson experiment (ATLAS, CMS), used data science to discover the definitive signature Higgs boson energy cascade, that was hypothesized by science, in the largest particle physics dataset in history. As in such experiments, foundation models[3], the currently most powerful data science models, are based on well-defined scientific domain models of natural language and vision.

---

[11] Most easily done with task-based problems since results expressed in tasks can be readily evaluated.

## 3. How data science got its name

The history of the emergence of data science provides insights as to why data science is referred to, incorrectly, as a science – a misnomer with significant negative consequences.

### 3.1. Tukey named his vision data analysis

In 1962[21], John Tukey, a prominent statistician, presented a prescient, laudable, vision to extend statistics from its confirmatory role that he saw as boring, formulaic, and rigid, confirming ever more precise answers to known quantities, e.g., inflation rate. He envisioned *exploratory data analysis[22]* with which to discover insights, versus answers, into "what lies beneath the data", ignoring obvious, recognizable patterns often sought in statistics, to seek insights that you never expected, always doubting your analysis and results. As a counterpoint to mathematics, mathematical statistics, and statistics that use theory rigidly to determine and validate what to derive from analyses, he saw data analysis, like science, requiring "art, flexibility, and judgement" to select analysis methods and interpret what is derived, based on "experience" (e.g., empirical results) guided, not determined, by theory. Based on mirroring the exploratory nature of scientific discovery, Tukey claimed what he called *data analysis* to be an experimental "science". Tukey's 1960's vision has expanded significantly by contributors from 40 disciplines, predominantly statistics, AI, and data systems with modeling, inference, workflows, visualization, and data mining. While AI algorithms (decision trees, learning algorithms using neural networks – Machine Learning, evolutionary, heuristic, and generative algorithms) had origins in the 1960s, they were not part of data science until the 1990s. Now, AI-based methods that qualitatively changed and now dominate data science in ways we are just beginning to understand.

By 2000, the power and applicability of the modern data science began to emerge far beyond Tukey's vision. In 2007[9], the brilliant Turing Award winner, Jim Gray, recognized the potential of what Gray called the "data-intensive science paradigm", and its primary application in eScience in which discoveries were made by computational means over Big Data. Gray gave a comprehensive view of computational data science as a new discovery paradigm, distinct from science, along the following evolution of research (knowledge discovery) paradigms.

- 1st paradigm: *Empiricism* (11th C BCE) Empirical evidence is discovered as observations from repeated experimentation as in agriculture to discover better ways to produce food in China and India.
- 2nd paradigm: *Science* (12th-19th C CE) Knowledge of the natural world is discovered empirically following the scientific method introduced by Roger Bacon (1219-1292), a polymath English monk, practiced intuitive versions of the scientific method in many domains - biology, optics, alchemy, and astronomy. The philosophical basis of science was defined by Francis Bacon (17th C CE) and Hume (18th C CE)[5]. Science was defined pragmatically in the 1830's by Comte[15].
- 3rd paradigm: Simulation (aka computational science, scientific computing, scientific computation) makes discoveries of natural and manmade systems by developing and experimenting with computational models that simulate those systems.
- 4th paradigm: Data-intensive science (aka Big Data analytics) makes discoveries by evidence gained from complex analytical methods, including AI methods, to extract insights from massive unstructured and structured datasets. Gray's insights are recognized in the history of science[18].

Gray distinguished the data-intensive, or data-driven, paradigm from its predecessors by analysis being the computational analysis of data. He identified scientific discovery to be the dominant application and predicted the emergence of eScience and cited as a leading example, particle physics research conducted using the Large Hadron Collider. Gray saw Big Data analytics as scientific not due to the amazing power and limitations, i.e., inscrutability, of data science methods, largely unknown in 2007; but due to his prediction of its dominate application in science that he called eScience. Science makes extensive use of data science and *vice versa*. Gray's contributions were to recognize "data-intensive" (Big Data) analysis, first, as fundamentally new, emerging *research paradigm* with massive potential worthy of recognition by the National Academies of Science[16] and second, as a means for realizing Tukey's exploratory data analysis vision.

### 3.2. Why data analysis was called *data science*

Tukey's compelling vision of exploratory data analysis was widely adopted, especially his insightful, exploratory objective. Tukey correctly observed that scientific discoveries are made through "exploratory analyses", "without its strict rules, guided, not determined by theory" using "art, flexibility, and judgement". These descriptions mirrored those made by Richard Feynman, the Nobel prize winning physicist and Tukey contemporary. For example, the Higgs boson was hypothesized in the mid-1960s. It was discovered empirically and collaboratively by thousands of physicists in two phases, the 45 year phase (1965-2010) that Tukey called exploratory science, and the 2 year (2010-2012) experimental phase when the scientific method and principles were followed rigorously. As a researcher, Tukey focused on the potential of exploration to discover "what lies beneath the data". As he did not apply data science in practice, he may have focused less on the experimental, confirmatory phase, i.e., conducting an analysis, where science and data science differ fundamentally. This inherent limitation of data science became more evident in the 2000s due to inscrutable AI-based data science entirely unforeseen in the 1960s. Tukey's vision of data analysis as an exploratory science may have led to the misnomer data science.

In 2015, Donoho, an influential statistician and student of Tukey, supported Tukey's claim that data analysis was a scientific endeavor[6][7] further confirming the *data science* misnomer. Tukey and Donoho based their claim of data analysis being a science on what they defined as essential characteristics of science.

- Intellectual content
- Organization into an understandable form
- "reliance upon the test of experience as the ultimate standard of validity" Tukey [21]
- "… to be entitled to use the word 'science' we must have a continually evolving, evidence-based approach." Donoho [6][7]

"Evidence-based approach" alludes to, but does not mention, empirical evidence obtained using the scientific method. The "test of experience" and "evidence-based approach" applies not just to science, but to interpretivism, e.g., history, archeology, economics, politics, and data science. The scientific version of the "test of experience" and "evidence-based approach" relate to the strict application of the scientific method in the empirical phase, including the evaluation under empirical conditions of causal hypotheses that cannot be done in data science or statistics both of which are on the analytical versus the domain side of problem solving. Finally, data science violates the definition of science and *vice versa* as described in the previous section, §2.

Donoho[6][7] makes the case for developing of a theory for "data science". "Fortunately, there is a solid case for some entity called 'Data Science' to be created, which would be a true science: facing essential questions of a lasting nature and using scientifically rigorous techniques to attack those questions". Donoho concludes "As data analysis and predictive modeling becomes an ever more widely distributed global enterprise, 'Science about Data Science' will grow dramatically in significance." The need to understand currently inscrutable AI-based data science methods is widely accepted. Unfortunately, no such theory, scientific or otherwise, has been proposed requiring, as an alternative, the development of demonstrable, standard AI solutions that do not prove but demonstrate plausible explanations and interpretations as in foundation models[3] that are predicted to be the future of applied data science. A theory of data science is an open research challenge. This research shares Donoho's research objectives quoted below from[6][7].

"GDS6: Science about Data Science. Tukey proposed that a 'science of data analysis' exists and should be recognized as among the most complicated of all sciences. He advocated the study of what data analysts 'in the wild' are doing and reminded us that the true effectiveness of a tool is related to the probability of deployment times the probability of effective results once deployed.

Data scientists are doing science about data science when they identify commonly occurring analysis/processing workflows, for example, using data about their frequency of occurrence in some scholarly or business domain; when they measure the effectiveness of standard workflows in terms of the human time, the computing resource, the analysis validity, or other performance metric, and when they uncover emergent phenomena in data science, for example, new patterns arising in data science workflows, or disturbing artifacts in published analysis results.

The scope here also includes foundational work to make future such science possible—such as encoding documentation of individual analyses and conclusions in a standard digital format for future harvesting and meta-analysis.

As data analysis and predictive modeling becomes an ever more widely distributed global enterprise, "science about data science" will grow dramatically in significance."

### 3.3. Why it matters

The "data science" misnomer reflects similar misnomers in the 18th and 19th centuries when Comte's belief in the value of science led to the scientific research paradigm being applied inappropriately to interpretivist disciplines. As in those times, the popularity of "data science" may have inappropriately led to its use in disciplines where it doesn't apply, where results must be scientific, possibly leading to the false belief that data science results are scientific, i.e., definitive. Data science can accelerate scientific discovery, however, the two employ distinct discovery methods from distinct paradigms.

Mislabeling a research discipline as a science is common possibly to confer rigor on those disciplines like social science and political science where few discoveries can made empirically since randomized control trials (RCTs) were infeasible until the advent of natural experiments that require naturally arising empirical conditions. The 2019 Prize in Economic Sciences was awarded for "the application of Natural Experiments applied to measure the effect of programmes and policies such as rural electrification and iron supplementation, and of uncontrolled events such as famines". In these cases, the Rwandan government applied different programs and policies in different regions of the country giving rise to natural experiments conducted across otherwise similar regions. Natural experiments are applied in social sciences[14] as data science.

The term *data science* implies a fundamental misunderstanding that data science' results are definitive thus limiting data science understanding and application. Such confusions can have serious consequences. First, understanding the true nature of data science facilitates identifying open challenges (e.g., defining an epistemology, recognizing its limitations and risks) and unfathomed opportunities. The power and widespread use of AI-based data science has led to world-wide calls to action to address its potential negative consequences and risks. This requires a deep understanding of data science, especially the currently inscrutable nature of AI-based data science. Second, defining the data science research paradigm, distinct from science, provides clear definitions of data science and data science problem solving to guide the development of data science disciplines and associated practical applications, just as in the 19th century the scientific paradigm guided the development of hundreds of scientific disciplines and myriad science-based industries. Third, as the data science research paradigm is distinct from science, it can be used within the scientific research paradigm, and *vice versa,* to address grand challenges, e.g., protein folding, to accelerate scientific discoveries that humans could not otherwise achieve. DeepMind has delivered spectacularly on its mission to accelerate scientific discovery for "solving intelligence to advance science and benefit humanity". Fourth, achieving discipline-specific data science solutions in one discipline can lead to expanding the scope of data science by generalizing the solutions to augment the data science paradigm (e.g., methods, models, principles, laws) that can be applied in other data science disciplines. Fifth, formal definitions of data science as a field of inquiry provides a coherent basis for the data science community in 40+ contributing disciplines, data science sub-areas, 100s of application domains, tens of thousands of applications, and industries to understand, use and evolve data science. Sixth, understanding data science as a science prevents understanding the potential of data science for addressing challenges and producing results that are beyond human capacity in scope, scale, and complexity using non-human, but inscrutable concepts, logic, and reasoning.

## 4. Data science multiple definitions challenge

This section addresses challenges within the data science multiple definitions challenge (§1.4) and intended solutions (§1.6) – the approaches taken to address them, the solution requirements and benefits, and the solutions themselves. As each case is analogous, a detailed description is given for the data science research paradigm case with subsequent cases described by their unique differences. In each case, a unified, coherent definition of a concept is developed from multiple accepted definitions for the concept using the paradigm, unifying framework, and semantic approaches and the unified definition of the data science research paradigm.

### 4.1. ... for the data science research paradigm

What is the data science research paradigm? How do you define a fundamentally new field of inquiry? This is done applying Comte's paradigms to understand the philosophy of data science – its nature, purpose, and methods – and to guide its definition using the data science reference framework, predominantly the ontology. First, apply the unifying framework approach, to identify the artifacts that constitute each component of the data science reference framework. Second, apply the paradigm approach to the myriad definitions of the artifacts to develop a unified understanding of each artifact. Third, from the unified understandings develop a unified, coherent definition of each artifact. Fourth, use the definitions of the constituent artifacts collectively to derive the properties, capabilities, and limitations of the data science paradigm, thereby defining the data science field of inquiry. Throughout, apply the semantic approach to ensure coherence and consistence with related definitions.

#### 4.1.1. Complicating factors

Consider the nature of the challenge in more detail. Data science has been emerging slowly since 1962 and rapidly since 2000, in over 40 disciplines and tens of thousands of applications, each resulting in largely independent definitions and terminology for data science artifacts and thereby for the data science research paradigm. Each of the $10^6$ data science publications – papers, reports, text books, course notes – define relevant data science artifacts. As data science is in its infancy, it is unlikely that they were developed using reference definitions. Despite application-specific inconsistencies or incompleteness, each definition is valuable for understanding and collectively defining the data science field of inquiry. While the multiple definitions challenge arises with as few as two data science artifact definitions, the scale of the challenge is enormous with many complicating factors of which a few are considered below.

First, consider the basic challenges of analyzing myriad definitions of data science artifacts to develop unified understandings and standard terminology. Precision required in this work as described in §1.3. Imprecise or complex terminology can impede unification. There is little challenge in understanding and unifying similar or identical definitions. Challenges arise for seemingly distinct definitions that reflect different aspects or perspectives of an artifact. If they accurately define distinct aspects of the same artifact, the definitions enrich its definition but require unification. For example, this research has discovered more than fifteen distinct, enriching definitions of *data science workflow*. More challenging is resolving inconsistent or conflicting definitions. Are they intended to define the same data science artifact? Are some incorrect to be ignored? Also challenging is to maintain the coherence and consistency of definitions. Unifying source definitions requires maintaining the semantic coherence of related artifact definitions. More complex again is to detect and maintain the coherence of artifacts beyond semantic relationships across a data science ontology and epistemology, e.g., the meaning of a system of computations. A full definition of data science requires defining the complete reference framework, i.e., axiology, ontology, epistemology, methodology, methods, and technology. This paper addresses artifacts in the ontology and some in the epistemology.

Second, due to data science being a very active in research and applications, data science artifact definitions are emerging and evolving rapidly. Such evolution is inherent in the infancy of an emerging field. The approaches and means are proposed to aid managing this evolution, e.g., detect and maintain semantic relationships with artifacts from source disciplines.

Third, AI-based data science is inscrutable, lacking *explanations* of data analyses solutions and *interpretations* of data science results that are neither reliable nor robust. Despite those limitations, it has been applied successfully to make major knowledge discoveries in many disciplines and applications. Inscrutability yields unanswered questions like what do deep learning models learn and how do they reason? Such knowledge would improve the understanding hence quality of data science designs, solutions and their explanations, and results and their interpretations. While these questions are unanswered, guidance for safe, trustworthy application is provided for foundation models[3] for important categories of data science problems and solutions. Demonstrable, standard data science solutions (§2.2) should be developed to demonstrate interpretations, and optionally explanations, meet acceptable requirements that pose acceptable risks when applied in practice. Unified, coherent, well-understood definitions of the data science artifacts contribute to addressing such challenges despite inscrutability. This emphasizes the importance, complexity, and depth of understanding required to develop unified, coherent definitions.

Fourth, Kuhn's *incommensurability thesis[13][20]*, that concerns multiple definitions of science, applies to all research paradigms including data science. Multiple definitions are a natural part of an emerging, evolving field of inquiry reflecting advances in the field as it evolves and matures. The thesis suggests that data science, data science disciplines, and data science results obtained under different definitions of data science may be incomparable without a means of translating between different definitions. In the worst case, one version of data science definitions may invalidate not only another version, but also invalidate its results. This may seem extreme, however, the challenge of determining the coherence of two versions of definitions of data science is complex. The proposed approaches are intended to identify and possibly resolve such inconsistencies including incommensurability.

Fifth, essential aspects of data science can be misunderstood. In addition to its inherent inscrutability, common misunderstandings of data science make it difficult to understand, define, and apply. Data science model parameters and datasets volumes can be beyond human capacity to understand in scope, scale, and complexity. Humans accustomed to reasoning in less than ten dimensions are incapable of reasoning in billions or trillions of dimensions or parameters. Intuition for such reasoning may come as data science matures over decades. Most human reasoning in logic, mathematics, science, and the humanities and in their corresponding education are directed at achieving certain outcomes. Consequently, most people, including Einstein by his own admission, are not comfortable with, educated or practiced in reasoning about uncertainty, counterfactual reasoning and uncertain, probabilistic or counterfactual outcomes. For over 50 years, Einstein considered the probabilistic nature of quantum mechanics to be counterintuitive, saying that "God does not play dice with the universe". Data science is exclusively about reasoning in uncertainty. The lack of comfort, education, and practice with uncertainty can lead to misunderstanding data science – its purpose, the nature of analyses, solutions, and results. A unified, coherent definition of data science must overcome these misunderstandings and may lead to a deeper understanding of data science, and perhaps of the world.

How do we understand and define the inscrutable – phenomena beyond human understanding? Consider data science as a field of inquiry and dark energy – the hypothesized dominant energy of the universe. Before these hypotheses, our world knowledge was adequate for life on the planet – AI-based data science was unknown before machine learning's rise in the 1990s, yet to be explained; dark energy was unknown before Hubble's 1929 Law, explained only in 1998. We think of such phenomena being defined by their constituents (data science artifacts; dark energy particles, fluids, and forces) and collectively by their unique properties, capabilities, and limitations. While inscrutable, they offer insights previously impossible and not otherwise possible – data science insights beyond human understanding in scope, scale, and complexity into phenomena; dark energy analogously into our universe. As with previous such challenges in computation, e.g., P=NP, and astrophysics, e.g., the nature and extent of the universe, inscrutability was reduced by persistent knowledge discovery. The proposed processes to understand and define data science may reduce the inscrutability of data science, or not!

### 4.1.2. Requirements and benefits of unified definitions

A unified, coherent definition of data science to be achieved with the above methods has several requirements. First, such definitions and terminology must be accepted by the data science community requiring that they reflect the multidisciplinarity of the community and scope of applicability of data science. Synonyms can be used for terminology required by sub-communities. Such acceptance is to be achieved by a refereed publication process of the proposed data science journal. Second, definitions should satisfy critical properties such as consistency and coherence with related definitions and desirable properties such as robustness and reliability. Third, specification means, i.e., graphs, schema, and language, must be able to represent essential properties and support definition, modification, and unification operations that aid in maintaining critical properties, as proposed for the online journal system[32]. Fourth, source definitions should be from the many ($10^6$) publications that meet requirements, e.g., acceptance using PaperRank, i.e., PageRank modified for this case. They should be curated for use, e.g., ensure that they are at the right level of abstraction, e.g., paradigm, category, discipline, or instance. Finally, a unified definition should subsume the essence of the candidate definitions and selectively include properties that may arise in individual source definitions but are applicable to all. To assist with the definition process and to illustrate their representation, 1,000 proof of concept, candidate definitions are provided [24]-[32]. A comprehensive definition of data science may take a decade or more. Consider the emergence and evolution of data science and the corresponding 400 year history of defining the scientific research paradigm made more complex by the inscrutability of AI-based data science and inevitable exploration, Tukey's passion, to discover what *lies beneath*.

There are many benefits of a unified, coherent definition of data science. First, a unified definition can be used to recognize data science as a *fundamental, independent research paradigm* distinct from other research paradigms including science. Second, such a definition is easier to understand and apply since it is simplified, less redundant, and more coherent than myriad definitions. A unified data science definition together with the philosophy of data science and data science reference framework provide a deeper understanding to guide its use and development. Third, standard definitions and terminology support understanding, collaboration, and cooperation across data science disciplines and those that employ data science – a persistent problem in science[8] and may help to identify and address incommensurability[20]. Data science's scope of applicability is inherent multi-disciplinary[12] requiring that it be defined to be amenable to all relevant domains, as observed by data science leaders, e.g., "A common intellectual framework can facilitate knowledge sharing about data science as a discipline across different the fields and domains using data science methods in their research"[19]. Fourth, this in turn provides a basis for the data science community to collaboratively evolve data science thus revealing data science's true scope of applicability. For example, the unification process enables concepts and techniques currently restricted to one data science discipline to be generalized up to the data science paradigm level then applied (down) to all relevant data science disciplines. This too has been recognized by data science research leaders (§1.5). Fifth, understanding what is understood helps to identify what has yet to be understood, e.g., a theory of data science, including a data science epistemology due to its inscrutable AI-based data science methods, solutions, results, and other open challenges. This benefit is based on the ancient maxim[13] that progress is made less by accumulating facts than by knowing what you do not know. Sixth, as data science becomes better understood, defined, taught, and practiced at all education levels, people will become more comfortable with it and better able to reason in uncertain and counterfactual worlds, hence better able to use data science to solve real world problems. Finally, defining the data science research paradigm (the first of six definitional challenges) can contribute to the addressing the remaining five challenges and *vice versa*.

The above requirements and benefits of a unified definition of the data science research paradigm apply to all six challenges. Only differences are described for the remaining five cases.

### 4.2. ... for data science disciplines

Data science artifacts have instances in hundreds of data science disciplines and tens of thousands of data science applications, hence there can be four orders of magnitude ($10^4$) more candidate definitions for cases that concern disciplines and data analyses, i.e., all but case 1. The vastly larger number of candidate definitions increases the multiple definitions challenge scope (disciplines) and scale (applications). In turn, the diversity of the increased scope and scale provide greater opportunities to enrich the resulting unified definitions.

What is a data science discipline? How is the data science research paradigm applied in all disciplines. How do you define the concept of a data science discipline? The goal here is to derive from accepted data science application definitions, a unified, coherent definition of a data science discipline using the paradigm and unifying framework approaches.

This case concerns the myriad definitions of data science applications across their vast scope and scale. The nature of this challenge differs from the above challenge for the data science research paradigm. A data science discipline is defined by applying the data science research paradigm to a specific category of phenomena. The definition of the data science discipline, i.e., concepts that define all data science disciplines, requires defining the generic nature of the datasets and their corresponding computational analytical methods and *vice versa* for all data science disciplines. These concepts are defined in the data science reference framework. Their definition in all data science disciplines involves details, such as requirements for techniques used to realize them. This uses the unifying framework approach to deduce from myriad versions of definitions, the generic properties for all data science disciplines.

---

[12] Data science is devoid of human concepts, including disciplines. Multidisciplinarity is a human concept concerning its scope of applicability.

[13] Posed by the fathers of natural philosophy and science – Thales, Anaximander and Anaximenes in 6[th] century BC.

Several requirements are unique to this case. First, the unified definition of data science must be adequate to guide the solution to this challenge. A second is to acquire and select those definitions that adequately represent all data science disciplines. This is challenging due to data science's currently unfathomed scope of applicability as seen in the continued rapid emergence of applications and disciplines. This requirement can be relaxed as data science is assumed to evolve perhaps for decades which in turn reinforces a third requirement that the unified definition be amenable to evolution maintaining coherence and consistency. Finally, the solution requires understanding the nature of domain knowledge and its role in defining all disciplines. A benefit of a unified definition of the concept of a data science discipline is its use to discover and define new data science disciplines. These requirements and benefits apply in all subsequent cases.

**4.3. … for a specific data science discipline**

How is data science applied in a specific discipline? How do you define a specific data science discipline? How is the data science research paradigm applied to a specific category of phenomena in a specific discipline? What considerations are required to specialize data science to a specific discipline, i.e., category of phenomena. How do data science disciplines differ? What can be learned addressing this challenge? The goal here is to derive from accepted data science application definitions, a unified, coherent definition of a specific data science discipline, e.g., are GPT-4 and Beijing Academy of AI Wu Dao 2.0 based on a common definition of the data science NLP discipline? If not, would such a definition be valuable? The paradigm approach applies in these cases.

These cases concern instances of multiple definitions of a specific data science discipline. Addressing this challenge has value if there are multiple definitions that are inconsistent or that address phenomena or analyses of different nature, scope, and scale, e.g., address different natural language problems or analyses, and if each definition contributes techniques to the discipline. The nature of this challenge is to apply the above factors, requirements, and benefits to specific data science discipline definitions using domain knowledge of the discipline. This involves applying the unified definition of data science disciplines to the specific discipline definitions considering the phenomena, datasets, and types of analysis specialized to those discipline definitions using domain knowledge of the discipline. In all such cases, the unified definition must be a specialization of the generic reference framework for the specific discipline. This case typically involves unifying small number of definitions of a specific data science discipline, e.g., in data science NLP publications. A benefit of such a unified definition is its use to refine and expand a specific data science discipline, e.g., would data science NLP be improved by unifying definitions from two distinct instances, e.g., statistical (finding statistical patterns), behavioral (finding behaviors in speech acts), and connectionist (combing both), or by unifying (determining inherent similarities) from NLP techniques, e.g., sentiment analysis, named entity recognition, summarization, topic modeling, text classification, keyword extraction, lemmatization and stemming?

**4.4. … for the concept of a data analysis**

What are data analyses? What is the data science problem solving paradigm category? How do you understand and define a fundamentally new problem solving paradigm category? The goal here is to derive from accepted definitions of data analyses in all data science applications, a unified, coherent definition of *data analyses* using two data science reference framework components – data science methodology (*the data science method, the data science workflow, governing principles*) and data science methods *(computational methods, their operands (datasets), data science solutions (trained methods for prepared datasets), results, governing principles)*. This uses the paradigm, unified framework, and semantic approaches.

This case concerns computational, algorithmic means for conducting data science problem solving to achieve the central purpose of data

science – data analyses. While the previous three cases involve concepts concerning states, this and subsequent cases involve computations. The challenges, approaches, requirements, and benefits of the following three cases are analogous to the previous cases modified by applying them to computations leading to significant complicating factors. This requires understanding computational and inferential thinking[1].

First, the data science methodology and methods components are seldom defined or specified for data analyses that are expressed as algorithms. Unification requires that the methodology and methods component definitions be defined, potentially deduced from applications using the paradigm, unifying framework, and semantic approaches. Fundamental differences between applications, e.g., image versus language recognition, are reflected in distinct specializations of the generic methodology (the data science method and workflow) and methods. The nature of computational methods are typically considered by analytical categories (e.g., regression, classification, factor, dispersion, discriminant, time series, decision trees) each of which can have multiple algorithmic implementations. Unification requires identifying the same, similar, or inconsistent computations of multiple data analyses and unifying similar computations. The process is challenging requiring extensive knowledge of computation and algorithms. Evidence of this difficulty can be seen in the many, often inconsistent, taxonomies of data science computational methods. Defining the nature of computations requires defining the nature of the datasets that they analyze and *vice versa*. Second, unification requires a common terminology. This is easy for data science artifacts since differences can be handled with synonyms. Finding common terminology for computational analyses is deeper than synonyms. Determining that multiple specifications define the same computation is as challenging as determining that two algorithms implement the same analysis. Third, defining the nature of data analyses requires defining the nature of the solutions produced and their explanations and of the data science results and their interpretations. The objective is to define their nature including the fact that AI-based data science solutions and results are inscrutable with techniques to achieve acceptable explanations and interpretations including their weaknesses and strengths, the risks of applying them in practice, and means for ameliorating the risks. These issues apply to the three cases for data analyses. Fourth, data science analyses, as defined in the methods component, are computations with computational properties, e.g., computability, reliability, and robustness. Requirements for such properties should be defined, e.g., AI-based data science methods are computable but are neither reliable nor robust.

### 4.5. ... for data analyses in all data science disciplines

What is data science problem solving as conducted in all data science disciplines? What is the generic data science problem solving paradigm category as conducted in all data science disciplines, i.e., specializing the computational problem solving category to all data science disciplines? The goal here is to derive from accepted definitions of data analyses in data science applications, a unified, coherent definition of the data science problem solving paradigm as conducted in all data science disciplines using the paradigm approach.

This case concerns how data analyses are conducted in all data science disciplines. It faces the same challenges as the above data analyses case, i.e., definitive properties are seldom defined above the algorithm level, hence, the nature of the data analyses conducted in all disciplines must be deduced from applications. For example, are there rules that define what computational analysis categories apply in all data science disciplines versus in specific data science disciplines? The unified definition of data analyses can be used to derive the nature of data analyses in all data science disciplines, e.g., properties of classifications that apply to all disciplines, i.e., all categories of phenomena.

### 4.6. ... for data analyses in a specific discipline

What is data science problem solving in a specific data science discipline? How is the data science problem solving paradigm category conducted in a specific data science discipline, i.e., applied to a specific category of phenomena? What considerations are required to specialize a data science discipline to a specific category of phenomena to define a specific data science discipline? How is the data science problem solving paradigm used to define (specialize) the data analysis concept to a specific data science discipline? How does the data science problem solving paradigm differ between data science disciplines? What can be learned by addressing this challenge? The goal here is to derive from accepted definitions of data analyses in a specific data science application category, i.e., applied to a specific class of phenomena, a unified, coherent definition of that data science discipline, e.g., data science NLP, using the paradigm approach.

This case concerns how data analyses are conducted in a specific data science discipline. It faces problems shared with other cases, i.e., the scope and scale problem, and the need to deduce definitive properties that are seldom defined. It involves specializing the definition of data analyses applied in all disciplines (§4.5) to a specific discipline. The benefit of this analysis is to improve or validate the definition of data analyses in a specific discipline by identifying and resolving differences across definitions with the benefit of correcting or extending the methodology and methods components for the specific data science discipline. For example, how do you define data science problem solving in NLP deduced from many relevant applications and their algorithms, e.g., support vector machines, Bayesian networks, maximum entropy, conditional random field, neural networks/deep learning? How are they similar or do they differ? How do you select an algorithm for a specific NLP problem? Another benefit is to use the unified definition of data analyses in a specific discipline to similarly expand the definition of data analyses as conducted in all disciplines (§4.5).

### 5. The data science reference framework

This work provides an understanding of data science as a field of inquiry based on classical paradigms. It defines challenges defining data science, approaches to develop solutions, means to define data science, and solution requirements and benefits. A classical definition of data science has six components – axiology, ontology, epistemology, methodology, methods, and technology. Such a comprehensive definition may take decades. The definitions pursued in this research aim for a minimal (necessary and sufficient) definition for understanding data science. The epistemology component, i.e., reasoning or theory, is an open research challenge and may remain so for some time.

The above results provide the basis developing a unified, coherent definition of data science based on the data science reference framework. Candidate definitions [24]-[31] provide proofs of concept and value of being implemented in an online system as a data science journal[32] for the data science community to review, define, evolve, authorize, and apply data science, possibly over a decade.

**Acknowledgement**

I thank Prof. John Mylopoulos, University of Toronto, for thoughtful comments on this paper.

**Data science reference framework**

**Appendix I: Reasoning in data science**

Many forms of reasoning are required to understand, define, and apply data science. The term reasoning is can include analyze, apply, define, deploy, design, determine, develop, discuss, ensure, evaluate, evolve, execute, govern, guide, identify, implement, interpret, operate, prove, refine, research, specify, teach, test, understand, verify, and more.

**Appendix II: Data science and science philosophies**

"Philosophy of science is a branch of philosophy concerned with the foundations, methods, and implications of science. The central questions of this study concern what qualifies as science (i.e., distinguish science from non-science) the reliability of scientific theories, and the ultimate purpose of science. This discipline overlaps with metaphysics, ontology, and epistemology, for example, when it explores the relationship between science and truth. Philosophy of science focuses on metaphysical, epistemic and semantic aspects of science. Ethical issues such as bioethics and scientific misconduct are often considered ethics or science studies rather than the philosophy of science." [Wikipedia]

Replacing *science* with *data science* gives a definition of the philosophy of data science that is discussed in this paper and is widely studied in the data science community.

"Philosophy of data science is a branch of philosophy concerned with the foundations, methods, and implications of data science. The central questions of this study concern what qualifies as data science (i.e., distinguish data science from non-data science) the reliability of data science theories, and the ultimate purpose of data science. This discipline overlaps with metaphysics, ontology, and epistemology, for example, when it explores the relationship between data science and truth. Philosophy of data science focuses on metaphysical, epistemic and semantic aspects of data science. Ethical issues such as bioethics and data science misconduct are often considered ethics or data science studies rather than the philosophy of data science."